\documentclass[runningheads]{llncs}
\usepackage[T1]{fontenc}
\usepackage{hyperref}
\usepackage{graphicx}
\usepackage{multirow}
\usepackage{multicol}
\usepackage{placeins}
\usepackage{xcolor}
\usepackage{subfig}

\newcommand{\ourMethod}{\texttt{LoGex}}

\usepackage{adjustbox}
\usepackage{amsmath}
\begin{document}
\title{\ourMethod: 
Improved tail detection of extremely rare histopathology classes via guided diffusion
}
\author{Maximilian Müller
\and
Matthias Hein
}

\authorrunning{Müller et al.}
\institute{University of Tübingen and Tübingen AI Center \\
\email{maximilian.mueller@wsii.uni-tuebingen.de}\\
}
\maketitle              %
\begin{abstract}
In realistic medical settings, the data are often inherently long-tailed, with most samples concentrated in a few classes and a long tail of rare classes, usually containing just a few samples. This distribution presents a significant challenge because rare conditions are critical to detect and difficult to classify due to limited data. In this paper, rather than attempting to classify rare classes, we aim to detect these as out-of-distribution data reliably. We leverage low-rank adaption (LoRA) and diffusion guidance to generate targeted synthetic data for the detection problem. We significantly improve the OOD detection performance on a challenging histopathological task with only ten samples per tail class without losing classification accuracy on the head classes.
\footnote{Code is available at \href{https://github.com/mueller-mp/logex}{github.com/mueller-mp/logex}}

\end{abstract}
\section{Introduction}
Limited data availability poses a significant challenge for medical image analysis, especially considering the critical need for safety and reliability. 
In practical scenarios, clinical datasets can be highly imbalanced \cite{holste2022longTailCXR,Zhou21ReviewMedImaging,DoesYourDerma2021}, with a significant portion of samples coming from a few classes (the head) and a large number of classes with only a few samples (the tail). 
While the primary focus is, e.g., on solving a classification task on the head effectively, 
ensuring that cases of rare or unknown diseases are not classified as one of the frequently occurring head classes is equally important.

Such setups can be framed as long-tailed classification tasks, where the goal is good classification performance on both head and tail. Several strategies have emerged to tackle class imbalances, typically involving targeted loss functions \cite{LdamKaidi2019,focal2020}, oversampling or reweighting \cite{cui2019classbalanced,Kang2020DecouplingCRT} or augmenting the dataset \cite{hemmat2023feedbackguidedEntropy}. However, some classes might be too infrequent to predict reliably \cite{Zhou21ReviewMedImaging}. Therefore, it can be helpful to switch to a more manageable task: Classifying the head classes and simultaneously \textit{detecting} tail samples as ''abnormal'' so that they can receive special treatment. This can, therefore, be framed as an OOD detection problem \cite{hendrycks2017MSP}, where the goal is good classification performance on an in-distribution (the head {classes}) and detection of samples that are out-of-distribution (i.e., do not belong to the in-distribution classes: {tail classes, but also potential new or unknown diseases}). 
When regarding the tail classes %
as out-distribution, OOD methods can be readily applied. Traditional approaches to OOD detection \cite{hendrycks2017MSP,hendrycks22Scaling,liu2020energy,LeeMahalanobis2018,RenRelMaha2021}
typically work under the assumption of no prior access to outlier instances, even though there are methods leveraging auxiliary data \cite{hendrycks2019oe,du2023dreamood}. 
Since, for the task at hand, one has specific knowledge about the out-distribution in the form of very few samples, this should be leveraged for improved detection performance. Except for \cite{DoesYourDerma2021}, who devised a hierarchical training strategy for detecting tail samples in a dermatological task, approaches that leverage OOD detection methods for tail detection
are, however, not well explored.

Simultaneously, the rapid development of publicly available, general-purpose diffusion models \cite{rombach2021highresolutionSD,HoDiffusionModelsPaper,Nichol21DiffusionModelsPaper2,song2021DiffusionModelPaper3} has led to numerous applications that aim at improved classification \cite{Ghalebikesabi2023DifferentiallyPD,Trabucco23EffDataAugDiff}. In medical tasks, diffusion models were predominantly used for synthesizing \cite{Franzes23DiffForMedicalImag,dorjsembe2022threedimensionalImageSyn,Yu23DiffBasedNucleiSeg,Frisch23RareCatGuidedDiff} and for enhancing  \cite{Yang23DiffMIC,Ma23DiffMedImageEnhancementPlugPlay,Shen23StainDiff} images in a variety of domains.
Within the scope of OOD detection in the medical domain, diffusion models have mainly been used for reconstruction-based methods \cite{graham2023unsupervisedOOD,Graham_2023_CVPRDDPPOOD,Mishra23DualCondOODdet}.
In other fields, diffusion models were successfully deployed for synthesizing data for imbalanced classification \cite{hemmat2023feedbackguidedEntropy} and for OOD detection \cite{du2023dreamood}, albeit without a notion of tail classes. 

In this work, we present \ourMethod{} 
(\underline{Lo}RA+\underline{G}uidance for improved detection of \allowbreak \underline{ex}tremely rare classes),
an approach tailored explicitly towards improved detection of tail samples in the case of extreme data scarcity {while maintaining good performance on the head classes}. 
Specifically, our contributions are:
\begin{itemize}
    \item We present an extremely challenging {long-tailed} histopathology task with only 10 samples per tail class
    \item We devise \ourMethod{}, combining LoRA finetuning and 
    guidance to synthesize tail samples that are useful for tail detection
    \item We report improved tail detection performance with \ourMethod{} compared to all baseline methods, without loss in head classification performance
\end{itemize}

\begin{figure}[htb]
    \centering
        \includegraphics[width=.99\textwidth]{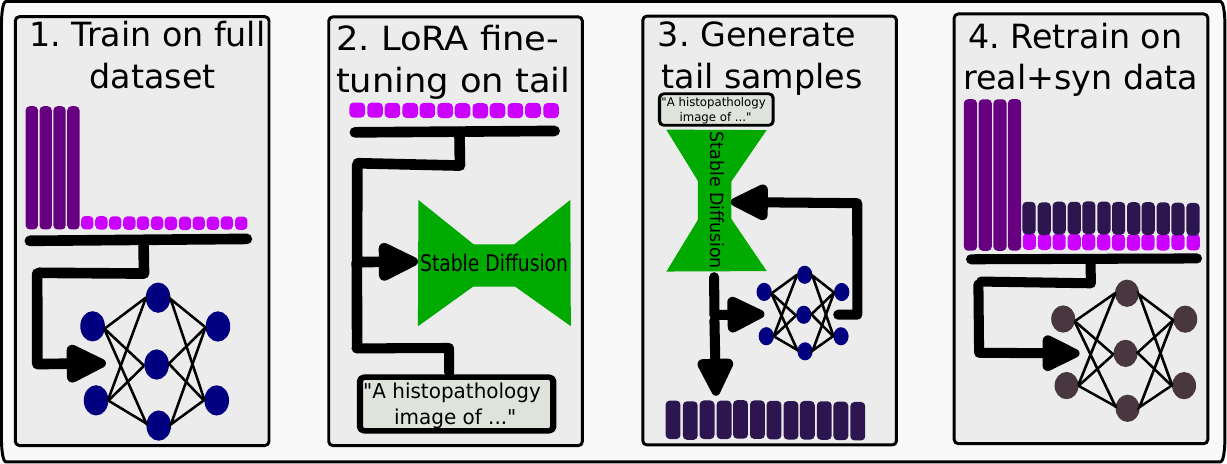}
    \caption{Workflow of \ourMethod: 1. We train an auxiliary classifier on the long-tailed dataset's head and tail classes. 2. We adapt a general-purpose diffusion model to the histopathology domain by applying LoRA finetuning \textit{only} on the tail samples. 3. We generate synthetic tail samples with the DiG-IN guidance from \cite{Augustin2023AnalyzingAE}. 4. We retrain a classifier by adding the synthetically generated tail samples to the train dataset.
    }
    \label{fig:flowchart}
\end{figure}

\section{Background: Diffusion models and Long-tailed classification}
\subsection{Diffusion models and DiG-IN guidance}
In score-based diffusion models, a noise sample is drawn from a prior distribution and iteratively denoised until a sample that is supposed to be from the desired data distribution is obtained. In latent diffusion models like StableDiffusion \cite{rombach2021highresolutionSD}, the denoising is performed for $T$ steps in the latent space of a variational autoencoder. 
The decoder of the VAE then transforms the final latent $z_0$ into pixel space. Deterministic solvers can perform the sampling in the latent space \cite{song2021DiffusionModelPaper3}, making the entire diffusion process deterministic and differentiable.
Additional conditioning signals $C$ can be employed in the diffusion process by sampling from a conditional distribution $p(z|C)$, where the conditioning $C$ can, e.g., be the encoding of a text prompt that the U-Net receives through cross-attention layers. While several approaches to guiding the diffusion process with other inputs exist  
\cite{song2021DiffusionModelPaper3,HoDiffusionModelsPaper},
the authors in \cite{Augustin2023AnalyzingAE} showed that fine-grained class structures can be best resolved when differentiating through the \textit{complete} denoising process \cite{wallace2023endtoend}. 
In their approach, called Dig-IN, the output of the diffusion model is taken as an input to a vision model (e.g. a classifier), and a loss $L$ is computed from the output of that vision model. The starting latent $z_T$ and the conditioning of the diffusion process are then optimized with respect to this loss by backpropagating \textit{through the entire diffusion process}, which is possible with deterministic solvers. This way, the diffusion process is guided to generate samples with low $L$. In our case, we will leverage DiG-IN to maximize the classifier-specific confidence of tail classes during the generation of synthetic tail samples.
\paragraph{LoRA.}
To leverage general-purpose large-scale diffusion models like Stable-Diffusion \cite{rombach2021highresolutionSD}, these models frequently must be adapted to specialized domains, e.g., medical ones.
While finetuning 
is often prohibitively expensive and difficult, \cite{hu2022lora} proposed to freeze the original weights instead and only to learn pairs of low-rank-decomposition matrices that are added to the attention layers (called LoRA). While initially designed for large language models, LoRA was shown to outperform other adaption methods \cite{hu2022lora} and to be effective for diffusion models, too. We will use this approach to adapt StableDiffusion to our task.

\subsection{Long-tailed classification and OOD detection}
There has been work directly linking long-tailed classification, OOD detection, and diffusion models. \cite{DoesYourDerma2021} devise a hierarchical outlier detection (HOD) loss function to improve the detection of rare and unseen classes of a dermatology classifier. This loss can be written as $\mathcal{L}_{HOD}=\mathcal{L}_{fine}+\lambda\mathcal{L}_{coarse}$, where $\mathcal{L}_{fine}$ is simply the cross-entropy loss over all classes, and the coarse-grained loss is $\mathcal{L}_{coarse}=-\sum_{c\in\{head,tail\}}\mathbf{1}(y=c)\log{p(c|\mathbf{x})}$ for an input sample $\mathbf{x}$. The term $p(c|\mathbf{x})$ is the head/tail probability, which is computed as the sum of the class probabilities for all classes in the head/tail: $p(c|\mathbf{x})=\sum_{k\in c} p(k|\mathbf{x})$ for $c \in \{\mathrm{head,tail}\}$.
At inference time $p(c=tail|\mathbf{x})$ is then used as OOD detection score. 

A promising approach leveraging diffusion models for synthesizing images for OOD detection is called Dream-OOD\cite{du2023dreamood}. In Dream-OOD a classifier is trained to learn a text-conditioned latent space of the in-distribution data. Then, outliers in the low-likelihood region of this latent space -- corresponding to the boundary of ID data -- are sampled and eventually decoded into pixel-space images by the diffusion model. 
A new classifier is then trained with cross-entropy on the in-distribution images, and a binary classifier is additionally employed on the model outputs to distinguish the in-distribution images from the synthetically generated samples.
{Another data generation approach is presented in \cite{hemmat2023feedbackguidedEntropy}, where the authors augment an imbalanced dataset with diffusion-generated synthetic images}. In particular, they suggest guiding the diffusion process by enforcing low entropy with classifier guidance. We refer to this approach as \textit{FG-entropy} (feedback-guided entropy) and include it in our experiments.

\section{Our method: \ourMethod} 
The high-level idea behind our method is that even though a diffusion model might not be able to generate realistic samples that are useful as additional training data to improve the classification performance, it might still be able to synthesize samples that are \emph{valid enough} to enhance OOD detection. To improve \textit{classification} on the tail classes, it is necessary to generate sufficiently diverse, class-specific samples from the tail distribution \cite{hemmat2023feedbackguidedEntropy}. For highly specialized domains like histopathology images, a generic diffusion model for natural images, e.g., Stable Diffusion \cite{rombach2021highresolutionSD} is unlikely to provide good sample quality directly, and even a specialized diffusion model for histopathology images should be unable to do this as for the rare tail classes it does not have enough training data. 
However, for OOD detection, the requirements for an auxiliary 'out-distribution' are lower: It is enough if the samples do \textit{not} belong to the in-distribution while still being informative for the task at hand. Therefore, we aim to generate samples with \textit{features} from the tail distribution and do not care if those samples are not entirely realistic, e.g., a mixture between two tail classes that might not occur naturally. While LoRA finetuning can help to adapt a pretrained diffusion model to a specialized domain, it is not enough to generate realistic enough tail features, as we show in Section~\ref{sec:ablation}. We, therefore, seek to include additional class-specific information in the diffusion process with DiG-IN guidance.
To this end, we devise \ourMethod, a strategy for the effective generation of samples with tail features. We illustrate our approach in Figure~\ref{fig:flowchart} and outline it in the following:

\begin{enumerate}
    \item \textbf{Train auxiliary classifier. } We train an auxiliary classifier on the long-tailed dataset (head and tail). This classifier will be used to guide the diffusion process and should thus ideally have some notion of what class-specific features both from the head and the tail classes look like. 
    \item \textbf{LoRA Finetuning. } To adapt a general-purpose diffusion model to the domain of histopathology images, we apply \textit{LoRA} finetuning with the tail samples to a pretrained general-purpose diffusion model. Since the finetuning process cannot resolve fine-grained class-specific differences, we purposefully exclude samples from the head for the finetuning process. In doing so, we attempt to avoid mixing up features from the head and tail as much as possible. For finetuning, we use the text prompt \textit{"A histopathological slide from a patient with }\texttt{<class>}\textit{"}. Details are reported in the Appendix, Table~\ref{tab:loradetails}.
    \item \textbf{Synthesize tail samples with DiG-IN guidance \cite{Augustin2023AnalyzingAE}. } We use the auxiliary classifier and the LoRA-finetuned diffusion model to perform the guided, class-specific generation of tail samples. In particular, we use the same prompt as employed during LoRA finetuning for each tail class. Additionally, during the diffusion process, we optimize the initial latent of the denoising process to maximize the confidence of the auxiliary classifier for the respective tail class.
\item \textbf{Augment the dataset and retrain the final classifier. } 
We augment the train dataset. For each tail class, we add the respective synthetically generated samples.
We then train a classifier with regular cross-entropy loss on the augmented train set and select the best-performing model on the validation set. As a selection criterion, we use the difference of the balanced accuracy on the head classes and the FPR of the tail detection. Since the generated synthetic images potentially contain features of different tail classes, the final classifier might confuse certain tail classes. For this reason, we use the tail probability (i.e., the sum of the individual tail class probabilities) as OOD score.
\end{enumerate}

\section{Experiments}
\subsection{Setup}
We create a challenging histopathology classification task based on the dataset of \cite{Kriegsmann23Skincancer}. The original dataset contains 129.364 image tiles from 386 cases manually annotated into 16 classes and split into a training, validation and test set. {The classification task on the original dataset can be solved very well} (97\% accuracy on the test set is achieved). To simulate a long-tailed scenario, 
we split and subsample the dataset into four head classes with more than 1000 training samples per class
and 12 tail classes with only ten samples per class. To make the task challenging, we designed the split so that similar classes are split into head and tail. 
We also subsample the available validation set to 100 samples per head class and ten samples per tail class. The test set we leave unchanged. Details on the dataset are illustrated in Figure~\ref{fig:distr} in the Appendix.
\paragraph{Experimental Details.}
We train a ResNet-50 for 60 epochs with AdamW and a cosine scheduler (base learning rate is $10^{-4}$). For the baseline long-tailed learning methods, we follow the setup of previous studies \cite{holste2022longTailCXR} and deploy three loss functions (standard cross-entropy, LDAM \cite{LdamKaidi2019}, and focal loss \cite{focal2020}). The latter two are specifically designed for highly imbalanced problems. We combine them with the reweighting strategy from \cite{cui2019classbalanced} (rw) and further evaluate classifier-retraining (CRT \cite{Kang2020DecouplingCRT}), deferred reweighting (DRW \cite{LdamKaidi2019}), and the HOD training strategy from \cite{DoesYourDerma2021}. As baselines using synthetic data, we adopt Dream-OOD \cite{du2023dreamood} and FG-Entropy \cite{hemmat2023feedbackguidedEntropy}. Since Dream-OOD does not explicitly use the available natural tail samples, we add them to the generated synthetic images, leading to a total out-distribution of size 1552.
Both for FG-Entropy from \cite{hemmat2023feedbackguidedEntropy} and \ourMethod{}, we generate 100 samples per tail class, as we found that increasing the number of synthetic samples beyond that did not bring additional gains (see App. Fig.~\ref{fig:nsamples}). For \ourMethod, we use StableDiffusion-1.4 as base model and perform LoRA on the cross-attention layers (details in App. Table~\ref{tab:loradetails}). We optimize the diffusion process until we reach a class confidence of at least $40\%$ on the auxiliary classifier. As ablation, we also evaluate \ourMethod{} without guidance, i.e., only after LoRA finetuning.

\subsection{Conventional OOD detection approaches}
\begin{table}[b]
    \centering
    \caption{\textbf{Comparing OOD scores and their tail-specific versions: }
    The accumulated tail probabilities $P(tail)$ perform best.
    }
    \begin{tabular}{lcccccc}
  & \text{MSP-head} & \text{MSP-tail} & \text{Maha-head} & \text{rMaha-head}& \text{Maha-tail}& \text{P(tail)} \\ \hline
\text{FPR}  & $30.98^{\pm2.36}$ & $25.33^{\pm1.91}$ & $53.40^{\pm3.10}$ & $32.44^{\pm4,06}$ &$98.46^{\pm0.63}$ & $\mathbf{25.03}^{\pm2.16}$ \\
\text{AUC} & $93.85^{\pm0.58}$ & $94.98^{\pm0.47}$ & $82.58^{\pm1.52}$ & $91.87^{\pm0.75}$ &$23.42^{\pm1.55}$& $\mathbf{95.01}^{\pm0.48}$ \\
\end{tabular}

    \label{tab:compareOODscores}
\end{table}
We first investigate if conventional OOD detection approaches 
can be effective in our setting. To this end, we evaluate commonly used OOD scores, like Max-Softmax (MSP) \cite{hendrycks2017MSP} and (relative \cite{RenRelMaha2021}) Mahalanobis distance \cite{LeeMahalanobis2018} for a ResNet-50 that was trained with cross-entropy and without synthetic data. The OOD scores are usually entirely based on the in-distribution (in our case, the head) and assume no knowledge about the out-distribution. Since we treat the tail as an out-distribution, we can also adopt the metrics to be tail-specific. E.g., the negative Max-softmax value across the tail classes (MSP-tail) is a natural OOD score for our setting. Similarly, Maha-tail denotes an OOD score based on the Mahalanobis distance to the tail distribution instead of the head. 
We report the results in Table~\ref{tab:compareOODscores}. 
The scores based on Mahalanobis distance
are outperformed by the MSP baselines and only the relative Mahalanobis distance (rMaha-head) performs comparable to the MSP baseline.
This is likely due to the low-data regime of our setup, which might make the mean and covariance estimation brittle, and underlines the need for tailored solutions. The best-performing method is based on the sum of the tail probabilities $P(tail)$, which was used in \cite{DoesYourDerma2021}. We will use it as the default OOD score for the main experiments. More (tail-specific) scores are reported in Table~\ref{tab:compareOODscoresall} in the Appendix.
\begin{table}[htb]
    \centering
    \caption{\ourMethod{} achieves the best tail detection performance and very high balanced accuracy on the head. We highlight the \textbf{best} and \underline{second best} methods.}

\begin{tabular}{lllccccc}
 & method & FPR & AUC & bAcc-head& bAcc-tail\\ \hline
\parbox[t]{3mm}{\multirow{9}{*}{\rotatebox[origin=c]{90}{without syn data}}} 
 &Cross-entropy  &$25.05^{\pm2.20}$ &$95.01^{\pm0.48}$ &$96.67^{\pm0.32}$&$44.47^{\pm4.68}$ \\
 &Cross-entropy + rw\cite{cui2019classbalanced}  &$25.86^{\pm3.73}$ &$94.16^{\pm0.76}$ &$94.02^{\pm0.73}$&$56.84^{\pm2.51}$\\
&Cross-entropy + HOD\cite{DoesYourDerma2021}  &$23.45^{\pm5.14}$ &$95.05^{\pm0.79}$ &$96.65^{\pm0.99}$&$49.75^{\pm2.41}$ \\
 &Cross-entropy + CRT\cite{Kang2020DecouplingCRT}  &$41.97^{\pm9.53}$ &$91.41^{\pm1.40}$ &$84.24^{\pm1.71}$&$68.26^{\pm1.91}$\\
 &Focal\cite{focal2020} &$23.48^{\pm6.32}$ &$94.74^{\pm1.20}$ &$96.27^{\pm0.91}$&$48.09^{\pm4.31}$ \\
  &Focal\cite{focal2020} + rw\cite{cui2019classbalanced}  &$25.23^{\pm6.02}$ &$93.56^{\pm1.19}$ &$94.02^{\pm1.30}$&$57.22^{\pm1.81}$\\
&LDAM\cite{LdamKaidi2019}  &$36.24^{\pm5.67}$ &$91.96^{\pm1.07}$ &$95.61^{\pm1.43}$&$56.62^{\pm2.14}$\\
 
 &LDAM\cite{LdamKaidi2019} + rw\cite{cui2019classbalanced}  &$50.08^{\pm6.54}$ &$89.95^{\pm1.06}$ &$91.21^{\pm1.27}$&$64.82^{\pm2.53}$\\
 &LDAM\cite{LdamKaidi2019} + rw\cite{cui2019classbalanced} + DRW\cite{LdamKaidi2019}  &$50.61^{\pm4.79}$ &$88.63^{\pm2.90}$ &$91.36^{\pm3.01}$&$59.60^{\pm18.43}$\\
 \hline
   \parbox[t]{3mm}{\multirow{4}{*}{\rotatebox[origin=c]{90}{syn data}}} 
 &Dreamood\cite{du2023dreamood}  &$26.21^{\pm2.75}$ &$92.33^{\pm0.68}$ &$\textbf{97.35}^{\pm0.89}$&--- \\
&FG-Entropy\cite{hemmat2023feedbackguidedEntropy}  &$20.66^{\pm4.29}$ &$\underline{95.48}^{\pm1.05}$ &$96.21^{\pm0.63}$&$52.04^{\pm4.72}$\\
&\ourMethod{} (only LoRA)  &$\underline{19.26}^{\pm3.22}$ &$\underline{95.48}^{\pm0.82}$ &$96.37^{\pm0.75}$&$49.42^{\pm5.01}$\\
&\ourMethod  &$\mathbf{16.25}^{\pm1.87}$ &$\mathbf{96.27}^{\pm0.33}$ &$\underline{96.97}^{\pm1.04}$&$54.80^{\pm1.27}$\\
\end{tabular}

    \label{tab:Overview}
\end{table}
\subsection{Main Results}

We report the results of our study in Table~\ref{tab:Overview}. 
Overall, \ourMethod{} clearly outperforms all other methods in terms of FPR and AUC while achieving very high balanced accuracy on the head classes. The second-best method is an ablation from \ourMethod{} with only LoRA and without DiG-IN guidance, highlighting that \textit{both} LoRA and guidance are crucial to achieve strong detection performance. The best baseline method without synthetic data is the hierarchical training strategy (HOD) from \cite{DoesYourDerma2021}, which achieves good detection performance and balanced head accuracy. Adding reweighting strategies (CRT, DRW, rw) improves tail classification performance for all three loss functions. This, however,  
comes at the cost of lower head accuracy and is also detrimental to the detection performance, and is therefore not a sensible strategy for our task.
For the reweighting methods, the accuracies on the tail classes are still significantly lower than on the head, underlining that classifying both head and tail is infeasible. Thus, tailored approaches like \ourMethod{}, which improve tail detection while performing strongly on the head classification task, are favourable. 
Dream-OOD achieves the best head accuracy but fails to detect tail samples effectively. The reason for this is likely the training scheme of Dream-OOD, which does not leverage the class structure of the available tail samples. On the contrary, FG-Entropy can leverage the class structure (i.e., trains the classifier on both head and tail classes) and is the best baseline method, but still outperformed by \ourMethod{} in both detection and classification tasks. FG-Entropy is designed for scenarios where the diffusion model is powerful enough to resolve fine-grained class differences by itself and the entropy guidance only acts as a regularizer for 
diversity. 
Due to the limited training data in our setup, the diffusion model (even after LoRA finetuning) cannot do so and needs class-specific guidance like in \ourMethod{} to create meaningful tail features. 
\begin{figure}[t]
    \centering
    \includegraphics[height=5cm]{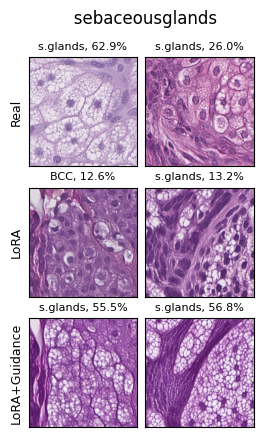}
     \includegraphics[height=5cm]{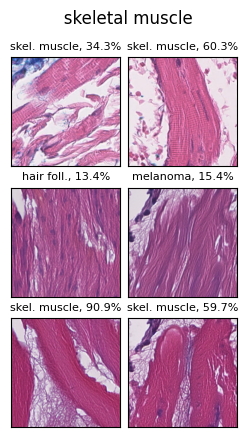}\includegraphics[height=5cm]{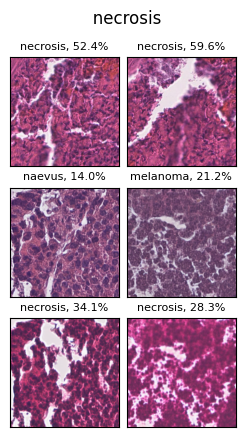}
    \includegraphics[height=5cm]{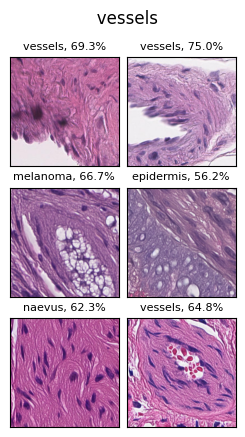}
    \caption{\textbf{Guidance matters: }We show samples from the train set (first row), samples generated with LoRA (second row), and samples generated with \ourMethod{} (third row) 
    and the predictions of a classifier trained on the original dataset.
    With \ourMethod{} the synthetic images are more often classified as the desired class.}
    \label{fig:samples}
\end{figure}
\subsection{Ablation: How good are the generated samples?}\label{sec:ablation}
To analyze the quality of the synthetic samples, we use a classifier that was trained on the original dataset from \cite{Kriegsmann23Skincancer}, achieving a tail accuracy of 96.1\%. {We only use the classifier for this retrospective analysis
and in no way during the main experiments.} On the synthetically generated tail samples we obtain an accuracy of $78.8\%$ for \ourMethod{}, 30.9\% for \ourMethod{} without guidance, and 35.6\% for FG-entropy, highlighting the improved quality of our synthesized images. 
We show samples for selected classes in Figure~\ref{fig:samples} and report the corresponding predictions. 
The samples generated with \ourMethod{} contain more pronounced features from the tail classes as compared to only LoRA finetuning, even though they are not always classified as the desired class. In a \textit{vessels} sample, for instance, the classifier detects features from \textit{naevus}. 
We note that such samples can still be useful for improved tail detection, albeit not for tail classification.
\vspace{-0.2cm}
\section{Conclusion}
We developed \ourMethod{}, an effective strategy that improves the tail detection performance without loss in classification performance of the head classes on a challenging histopathological task. We combine targeted classifier guidance and LoRA finetuning to adapt a pretrained general-purpose diffusion model to synthesize useful samples, outperforming other methods by a clear margin.
\FloatBarrier
\begin{credits}\subsubsection{\ackname} 
We acknowledge support from 
the DFG (EXC number 2064/1, Project number 390727645)
and 
the Carl Zeiss Foundation
in the project 
"Certification and Foundations of Safe Machine Learning Systems in Healthcare".
\subsubsection{\discintname}
The authors have no competing interests to declare.
\end{credits}
\bibliographystyle{splncs04}
\bibliography{main}

\newpage
\FloatBarrier
\appendix
\section*{Supplementary Material}

\begin{table}[htb]
    \centering
    \caption{\textbf{Comparing OOD scores: }
    Full version of Table~\ref{tab:compareOODscores} with more OOD scores and their tail-specific version. $P(tail)$ achieves the best performance.  
    }
    \begin{tabular}{l|c|c}
Score & FPR & AUC\\ \hline
P(tail)&$\mathbf{25.03}^{\pm2.16}$ &$\mathbf{95.01}^{\pm0.48}$\\
MSP \cite{hendrycks2017MSP} head &$30.98^{\pm2.36}$ &$93.85^{\pm0.58}$\\
 MSP \cite{hendrycks2017MSP} tail &$25.33^{\pm1.91}$ &$94.98^{\pm0.47}$\\
Energy \cite{liu2020energy} head &$39.17^{\pm5.30}$ &$91.79^{\pm0.51}$\\
 Energy \cite{liu2020energy} tail &$89.44^{\pm6.42}$ &$63.90^{\pm2.54}$\\
MaxLogit \cite{hendrycks22Scaling} head &$39.02^{\pm5.27}$ &$91.79^{\pm0.50}$\\
 MaxLogit \cite{hendrycks22Scaling} tail &$89.37^{\pm6.47}$ &$63.77^{\pm2.49}$\\
 Maha dist. \cite{LeeMahalanobis2018} to head& $53.41^{\pm3.17}$ & $82.58^{\pm1.52}$\\
Maha dist. \cite{LeeMahalanobis2018} to tail & $98.46^{\pm0.63}$ & $23.42^{\pm1.55}$\\
Maha dist. \cite{LeeMahalanobis2018} to tail - to head &$34.72^{\pm{5.11}}$&$89.23^{\pm{1.28}}$\\
relative Maha distance  \cite{RenRelMaha2021}& $32.45^{\pm4.06}$ & $91.88^{\pm0.75}$\\
\end{tabular}

    \label{tab:compareOODscoresall}
\end{table}

\setlength{\tabcolsep}{5pt}
\begin{table}[htb]
    \centering
    \caption{\textbf{LoRA training details.}
    }
    \begin{tabular}{lc}
rank & 4\\
learning rate &1e-04\\
train steps &15000\\
lr scheduler&cosine\\
resolution&512 \\
augmentation& center-crop and random-flip
\end{tabular}

    \label{tab:loradetails}
\end{table}
\setlength{\tabcolsep}{2pt}

\begin{figure}[htb]
    \centering
    \includegraphics[width=.7\textwidth]{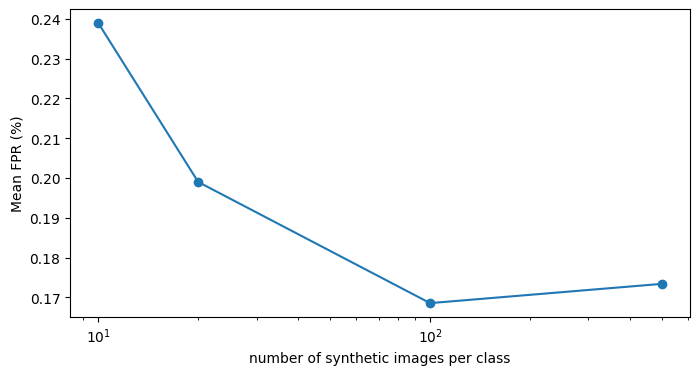}
    \caption{Ablation on the number of synthetic images per tail class: Adding more than 100 samples leads to a slight increase in FPR, but still outperforms baselines. We hypothesize that the relative importance of the natural tail samples decreases when too many synthetic images are used.}
    \label{fig:nsamples}
\end{figure}

\begin{figure}[htb]
    \centering
    \includegraphics[width=.7\textwidth]{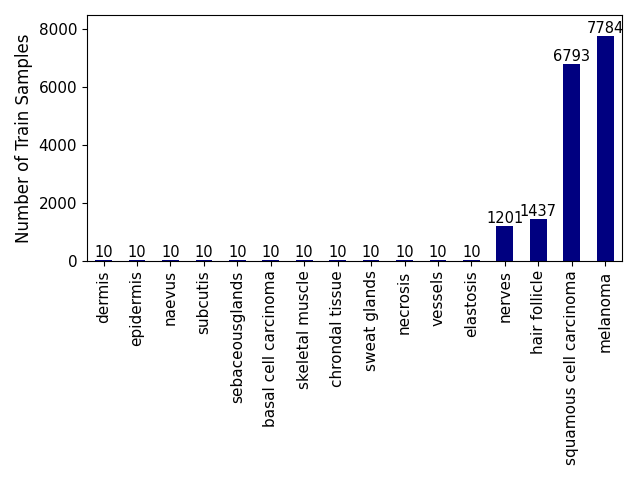}
    \caption{Class distribution of the train dataset.}
    \label{fig:distr}
\end{figure}

\begin{figure}[htb]
    \centering
    \includegraphics[height=5cm]{figures/samples/sebaceousglands-samples-offset0.png}
     \includegraphics[height=5cm]{figures/samples/skeletalmuscle-samples-offset0-target.png}\includegraphics[height=5cm]{figures/samples/necrosis-samples-offset0-target.png}
        \includegraphics[height=5cm]{figures/samples/vessels-samples-offset0-target-newm.png}

        \includegraphics[height=5cm]{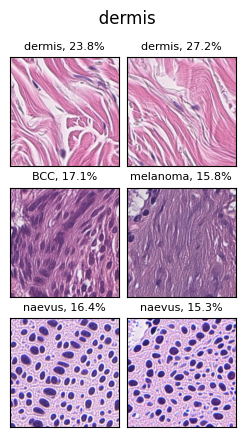}
     \includegraphics[height=5cm]{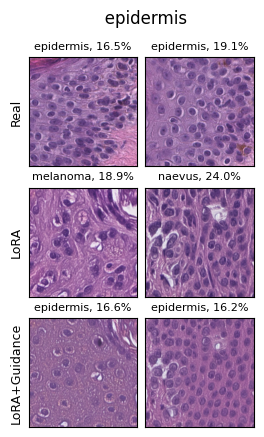}\includegraphics[height=5cm]{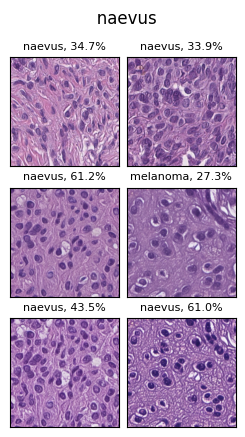}
    \includegraphics[height=5cm]{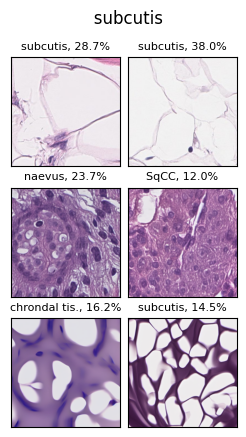}

            \includegraphics[height=5cm]{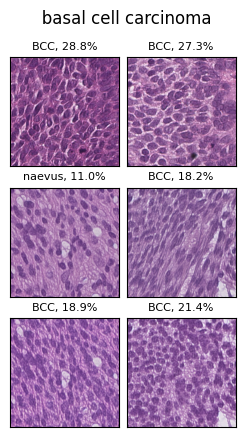}
     \includegraphics[height=5cm]{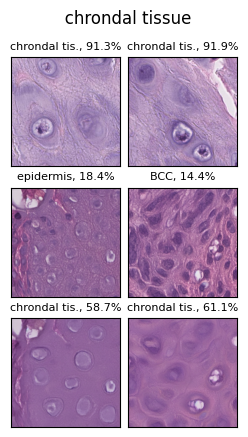}\includegraphics[height=5cm]{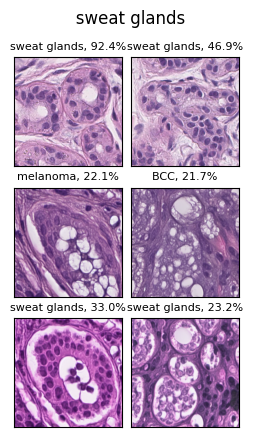}
    \includegraphics[height=5cm]{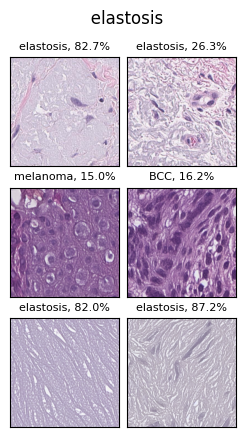}

    \caption{\textbf{Samples from all classes: }We show samples from the train dataset (first row), samples generated with LoRA (second row), and generated with \ourMethod{} (third row). For each sample, we report the corresponding prediction of a classifier trained on the original dataset from \cite{Kriegsmann23Skincancer} (achieving a tail accuracy of 96.1\%). With \ourMethod{} the synthetic images are more often classified as the desired class.}
    \label{fig:samples-all}
\end{figure}
\end{document}